\documentclass[journal,12pt,onecolumn,draftclsnofoot]{IEEEtran}

\usepackage{cite, amsmath, comment, xspace, color, tabu, algorithm, algpseudocode, amssymb, multirow}
\usepackage{bm} 
\usepackage{amsthm}
\usepackage{graphicx}
\usepackage{subfigure}
\usepackage{array}
\usepackage{bbm} 
\usepackage{dsfont} 
\usepackage{mathtools} 

\usepackage{diagbox}
\usepackage{enumitem}

\usepackage{tcolorbox}
\tcbuselibrary{breakable}
\newtcolorbox{textbox}{
    sharp corners,
    colback=white,
    colframe=darkgreen,
    boxrule=.5pt,
    breakable,
}
\newtcolorbox{textbox1}{
    sharp corners,
    colback=white,
    colframe=red,
    boxrule=.5pt,
    breakable,
}

\newtheorem{assumption}{Assumption}

\newtheorem{theorem}{Theorem}[section]

\newtheorem{property}{Property}

\usepackage{mathtools}

\definecolor{darkgreen}{rgb}{0.0, 0.5, 0.0}

\begin{document}
\title{A Metric Topology of Deep Learning for Data Classification}
\author{Jwo-Yuh Wu$^\dagger$, Liang-Chi Huang$^\dagger$, Wen-Hsuan Li$^\dagger$, and Chun-Hung Liu$^\ddagger$
\thanks{\!\!\!\!\!\!$^\dagger$ Institute of Communications Engineering, and the College of Electrical Engineering, National Yang Ming Chiao Tung University, Taiwan (jywu@nycu.edu.tw, lchuang@nycu.edu.tw, vincent@nycu.edu.tw).
\\$^\ddagger$ Department of Electrical and Computer Engineering, Mississippi State University (chliu@ece.msstate.edu)}
}
\maketitle

\vspace{-1.5cm}

\begin{abstract}
Empirically, Deep Learning (DL) has demonstrated unprecedented success in practical applications. However, DL remains by and large a mysterious “black-box”, spurring recent theoretical research to build its mathematical foundations. In this paper, we investigate DL for data classification through the prism of metric topology. Considering that conventional Euclidean metric over the network parameter space typically fails to discriminate DL networks according to their classification outcomes, we propose from a probabilistic point of view a meaningful distance measure, whereby DL networks yielding similar classification performances are close. The proposed distance measure defines such an equivalent relation among network parameter vectors that networks performing equally well belong to the same equivalent class. Interestingly, our proposed distance measure can provably serve as a metric on the quotient set modulo the equivalent relation. Then, under quite mild conditions it is shown that, apart from a vanishingly small subset of networks likely to predict non-unique labels, our proposed metric space is compact, and coincides with the well-known quotient topological space. Our study contributes to fundamental understanding of DL, and opens up new ways of studying DL using fruitful metric space theory.
\end{abstract}

\section{Introduction}
\subsection{Background and paper contributions}
Deep learning (DL) is epoch-making and its study grows apace in the recent years \cite{Goodfellow16,Higham19}. Despite phenomenal empirical performances having been demonstrated in many application domains, less is known about the theoretical side. Investigation into \textit{mathematics of} DL is therefore underway, from training/architecture optimization \cite{Bolcekei19, Cacciola23,Elad23,Soltanolkotabi19}, generalization error analysis \cite{Grohs23,Mohri18,Shwartz14,Wiatowski18}, to interplay with quantum physics \cite{Dunjko18,Hush17,Schuld15}, towards building mathematical foundations underlying its sheer success. Our study in this paper is in pursuit of fundamental understanding of DL tasked with data classification, and is well motivated by the toy example given below. Consider a one-layer DL network consisting of six real parameters $(w^{(1)},w^{(2)},w^{(3)},w^{(4)},b^{(1)},b^{(2)})$ responsible for binary classification over $\mathbb{R}^2$, as depicted in Fig. 1; three distinct network parameter vectors $\{\widetilde{w}_1,\widetilde{w}_2,\widetilde{w}_3\}$, along with their pair-wise Euclidean distances, are then listed in Table \ref{tab:toy}. For randomly sampled testing points, the labels output by $\{\widetilde{w}_1,\widetilde{w}_2,\widetilde{w}_3\}$ are shown in Fig. 2, in which close network parameter vectors (say, $\widetilde{w}_1$ and $\widetilde{w}_2$ because $\|\widetilde{w}_1-\widetilde{w}_2\|_2=0.283$, less than $\|\widetilde{w}_1-\widetilde{w}_3\|_2=3.959$ and $\|\widetilde{w}_2-\widetilde{w}_3\|_2=4.232$) are seen to perform nonetheless drastically different. Obviously, the Euclidean metric fails to discriminate networks according to their classification outcomes. This then inspires construction of a meaningful metric, against which DL networks yielding similar data classification performance are close. Not only can such a metric (if available) offer accurate performance assessment, but also give birth to a potential metric topology underlying DL for data classification, enriching theoretical research of DL with fruitful metric space theory \cite{Heinonen01,Marsden93,Rudin23}.

\begin{figure}[t!]
  \centering
  \label{fig_1}\includegraphics[width=0.92\columnwidth]{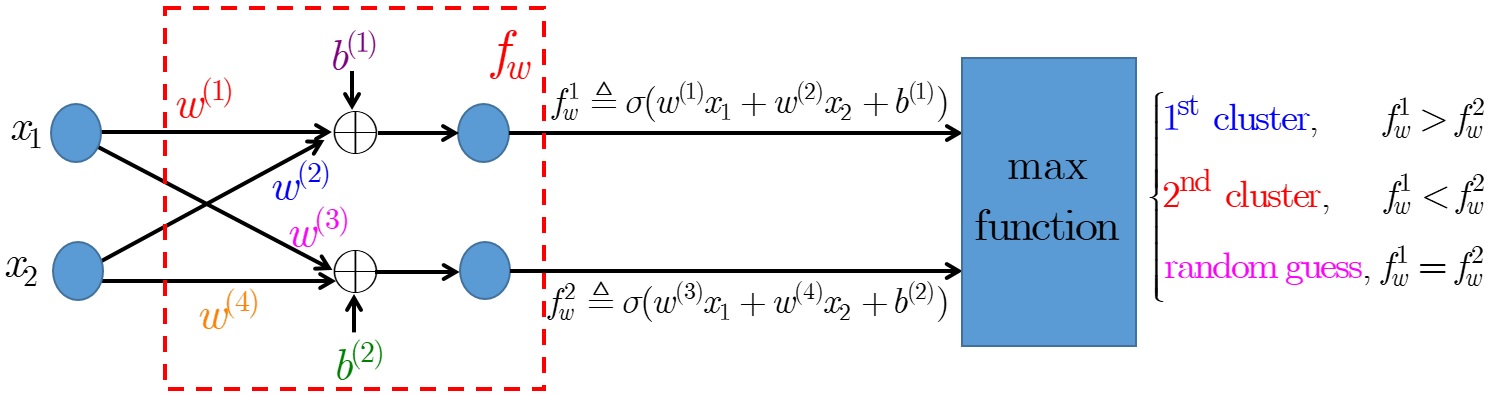}
  \vspace{-1.7em}
  \caption{Illustration of a DL network with one hidden layer, characterized by six real parameters $w=(w^{(1)},w^{(2)},w^{(3)},w^{(4)},b^{(1)},b^{(2)})$, to conduct binary classification for input data from $\mathbb{R}^2$. The input-output relation of the network is represented by a function $f_w:\mathbb{R}^2\to\mathbb{R}^2$, with $f_w^j$ the $j$th component. The network labels a test data point $x=(x_1,x_2)\in\mathbb{R}^2$ as $i$ if $f_w^i(x)>f_w^j(x)$, and makes a random guess if  $f_w^1(x)=f_w^2(x)$.}
\end{figure}

\begin{table}[t!]
\footnotesize
\caption{Network parameter vectors and their pair-wise Euclidean distances.}\label{tab:toy}
\vspace{-2em}
\begin{center}
  \begin{tabular}{|c|c|c|c|c|c|} \hline
   $\widetilde{w}_1$ & $\widetilde{w}_2$ &$\widetilde{w}_3$ &$\|\widetilde{w}_1-\widetilde{w}_2\|_2$ &$\|\widetilde{w}_1-\widetilde{w}_3\|_2$ &$\|\widetilde{w}_2-\widetilde{w}_3\|_2$ \\ \hline
   $(0.8,1,1,1,0.9,1)$& $(1,1,1,1,1.1,1)$ & $(-2,1,1,1,-1.9,1)$ & 0.283 & 3.959 & 4.243 \\ \hline
  \end{tabular}
\end{center}
\end{table}

\begin{figure}[t!]
  \centering
  \label{fig_2}\includegraphics[width=0.92\columnwidth]{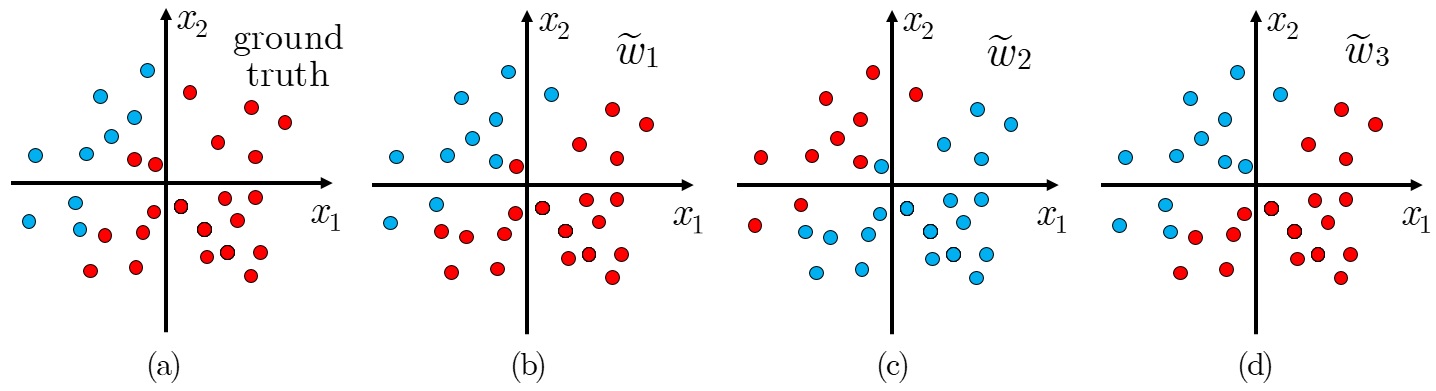}
  \vspace{-2em}
  \caption{An illustration of classification outcomes of three networks (parameter vectors listed in Table \ref{tab:toy}) for randomly sampled testing points in $\mathbb{R}^2$, where the ground truth is shown in (a). Sub-figures (b), (c), and (d) are the labeled results corresponding to parameter vectors $\widetilde{w}_1$, $\widetilde{w}_2$, and $\widetilde{w}_3$, respectively.}
\end{figure}

  In light of the above points, the main contributions of this paper can be summarized as follows.
\begin{enumerate}
\item We propose a new distance measure from a probabilistic point of view, against which DL networks delivering similar classification outcomes are close. Conceptually, our proposed solution gauges the likelihood that the decisions of two networks disagree; in particular, the distance from a given DL network to the ground truth classifier turns out to be its generalization error \cite{Bousquet02,Mohri18,Shwartz14,Zhang21}.
\item We show the distance measure defines such an \textit{equivalent relation} that networks yielding identical performances belong to the same equivalent class. The associated \textit{quotient set modulo the equivalent relation}, therefore, is composed of disjoint equivalent classes, each one consisting of DL networks performing equally well.
\item Importantly, our distance measure can provably serve as a metric on the quotient set. Under quite mild assumptions on the input data distribution it is shown that, apart from a vanishingly small subset of networks likely to predict non-unique labels, the proposed metric space is compact.
\item Finally, it is shown that the proposed metric topology on the quotient set coincides with the well-known quotient topology \cite{Gamelin99}, namely, the collection of all subsets of the quotient set whose inverse images under the projection map \cite{Munkres00}, in our case from the Euclidean network parameter space into the proposed metric space, are open.
\end{enumerate}

\subsection{Related works}
The study of DL from a topological perspective is of great relevance and received considerable attention of late \cite{Bendich16,Lee11,Singh07,Wu24}. This is because, above all, a DL network is essentially a composition of continuous functions, and topology is concerned chiefly about continuous functions, the spaces being acted upon, and structural properties preserved. Additionally, topology can characterize intrinsic geometric features of large and complex datasets \cite{Carlsson09,Krim16,Xia17} that are very helpful to guide learning with big data. Existing related works can be categorized into the following three topics (some selected publications listed):
\begin{enumerate} 
\item \textit{Topological features of data and networks}: Identification of topological invariants, or the likes, of complex datasets is addressed in \cite{Adams17, Bubenik15} to gain insights into the geometry of data. Evolution of the input geometric feature layer after layer is examined in \cite{Naitzat20}, in order to clarify the possible induced deformation when data piping through the network. In addition, evolution of the network weights as a set of points of cloud during training is examined in \cite{Gabella21} from a topological point of view. Related applications in material science and financial market analysis can be found in \cite{Gidea20,Hiraoka16}.
\item \textit{Topological signature extraction}: 
Learning a task-optimal representation of the topological signatures and a related input-layer design formulation is addressed in \cite{Hofer17}. Extension by using additional kernel methods for better encoding the learned feature is obtained in \cite{Carrière17,Reininghaus15}. Moreover, structured output spaces with controlled connectivity have been applied in autoencoder design to preserve essential topological properties \cite{Hofer19}. Extraction of topological structures, embedded in the order of text sequences, for enhancing text classification is addressed in \cite{Gholizadeh18}.

\item \textit{Topology-aided training}:  
Learning guided by topological priors towards better image segmentation is discussed in \cite{Clough22,Hu19}. By exploiting certain topological structures of convolutional networks, fast and efficient network training can be achieved \cite{Love23}. Design of topological penalty for the training of classifiers in order to simplify decision boundaries in addressed in \cite{Chen19}. For fake news detection in natural language processing, incorporation of topological features is shown to improve training accuracy with limited data size \cite{Deng22}.
\end{enumerate}

All the aforementioned works focus on topological feature characterization and extraction, emphasizing on applications in network training and architecture design. Methodologically, they are capitalized on persistent homology, an algebraic-topology tool introduced in the recent advances of topological data analysis \cite{Aktas19, Barbarossa20,Otter17}; a detailed survey of such topological deep learning can be found in \cite{Edelsbrunner22,Wasserman18,Zia24}. By sharp contrast, our study in this paper aims to assess accurate classification performance gap between DL networks, whereby a new metric topology of DL for data classification is developed and analyzed. To the best of our knowledge, formulation and analysis of DL under the metric topology framework remain lacking in the vast literature. Our study is the first attempt to fill this void, and presents new mathematical aspects of DL for data classification.

The rest of this paper is organized as follows. Section \ref{sec_2} first briefly goes through the network model. Section \ref{sec_3} introduces the proposed distance measure. Section \ref{sec_4} then develops the proposed metric topology. Section \ref{sec_5} goes on to present key mathematical properties of the metric topology. Finally, Sections \ref{sec_6} concludes this paper and discusses some future research directions. To ease reading, detailed mathematical proofs and derivations are relegated to Appendix.

\subsection{Notation list}
For a Lebesgue measurable subset $S$ of the Euclidean space, the symbol $|S|$ represents its measure; if $S$ is finite, $\text{card}(S)$ its cardinality. For a real matrix $A\in\mathbb{R}^{m\times n}$, $\text{vec}(A)\in \mathbb{R}^{mn}$  is obtained by stacking the columns of $A$ consecutively \cite{Macedo13}. $\mathbb{R}^+_0\subset \mathbb{R}$ is the subset of non-negative real numbers. $\|\cdot\|_2$ denotes the Euclidean two-norm.

\section{Network model}\label{sec_2}
We consider an $L$-hidden-layer DL network, with input domain $\Omega\subset\mathbb{R}^{n_0}$ and output space $\mathbb{R}^{K}$, whose purpose is to classify a given test data point $x\in\Omega$ into one of $K$ target classes; here we focus on the supervised learning case so that $K$ is known in advance. At the $l$-th layer, $1\leq l\leq L+1$ (the $L+1$th layer is the output layer), the output $y\in\mathbb{R}^{n_{l-1}}$ from the $(l-1)$-th layer is mapped into $\mathbb{R}^{n_{l}}$ according to
\begin{equation}
\label{eq:2_1}
g_l(y)=\sigma(W_l y+b_l), 1\leq l\leq L+1,
\end{equation}
in which $W_l\in\mathbb{R}^{n_{l}\times n_{l-1}}$ and $b_l\in\mathbb{R}^{n_{l}}$ are, respectively, the weight matrix and the bias vector, and $\sigma:\mathbb{R}^{n_{l}}\to\mathbb{R}^{n_{l}}$ is the element-wise activation function, each one assumed to be continuous and strictly increasing. Towards a compact description of the network, we stack the weight and bias $\{W_l,b_l\}$ of the $l$-th layer into
\begin{equation}
\label{eq:2_2}
w_l\triangleq \big(\text{vec}(W_l),b_l\big)\in\mathbb{R}^{n_l(n_{l-1}+1)},  1\leq l\leq L,
\end{equation}
concatenated one next to another to obtain the network parameter vector
\begin{equation}
\label{eq:2_3}
w\triangleq  (w_1,w_2,\dots,w_L,w_{L+1})\in\mathbb{R}^{m}, \text{where } m\triangleq \sum_{l=1}^{L+1}n_l(n_{l-1}+1).
\end{equation}
In the sequel we denote by $\mathcal{W}\subset\mathbb{R}^m$ the collection of all network parameter vectors. The input-output relation of a DL network, represented by $w\in \mathcal{W}$, can be described by a continuous function $f_w:\Omega\to\mathbb{R}^K$ as
\begin{equation}
\label{eq:2_4}
f_w\triangleq g_{L+1}\circ g_{L}\circ g_{L-1}\circ \dots \circ g_{1}, 
\end{equation}
which is a composition of the mapping \eqref{eq:2_1} layer after layer. Let $f_w^j:\Omega\to\mathbb{R}$  be the $j$th component of $f_w$, and $y_w(x)\subset \{1,\dots,K \}$ the label of the data point $x\in\Omega$ predicted by $f_w$. Then we have
\begin{equation}
\label{eq:2_5}
y_w(x)= \underset{1\leq j\leq K}{\arg\max}\text{ }f^j_w(x). 
\end{equation}
If $y_w(x)$ is not a singleton, the network makes a random guess among the reported indexes. A schematic summary of the network model for data classification is given in Fig. 3. The following assumption is made throughput the paper.
\begin{assumption}\label{as:measure}
The input domain $\Omega\subset\mathbb{R}^{n_0}$ is endowed with a probability measure $\mu$ so that $(\Omega, \mathcal{A}, \mu)$ is a probability space, where the event space $\mathcal{A}$ is a sigma algebra of sunsets of $\Omega$, and the probability measure of $E\in\mathcal{A}$ is denoted by $\mu(E)$. 
\end{assumption}

\begin{figure}[t!]
  \centering
  \label{fig_3}\includegraphics[width=0.95\columnwidth]{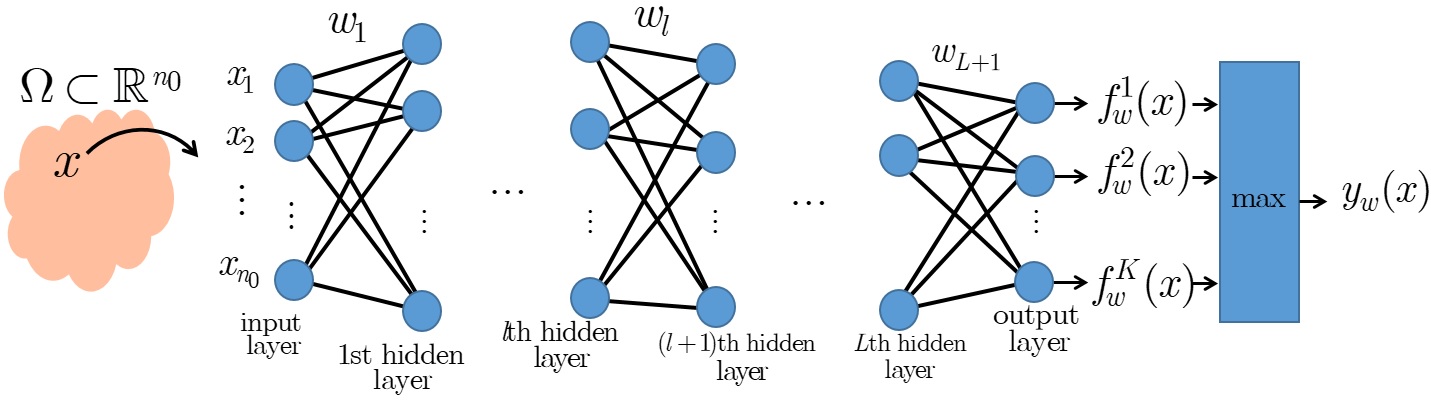}
    \vspace{-1.7em}
  \caption{An illustration of the input-output relation of a DL network, with $L$ hidden layers, for data classification.}
\end{figure}

\section{Proposed distance measure}
\label{sec_3}
\subsection{Formulation}
\label{sec:3_1}
Construction of our proposed distance measure is built on an equivalent characterization of the decision rule \eqref{eq:2_5}. For definiteness, we decompose the output space $\mathbb{R}^K$ into
\begin{equation}
\label{eq:3_1}
\mathbb{R}^K=\mathcal{R}_0\bigcup_{k=1}^{K}\mathcal{R}_k,
\end{equation}
in which
\begin{equation}
\label{eq:3_2}
\mathcal{R}_k\triangleq\{z=(z_1,\dots, z_K)\in\mathbb{R}^K\mid z_k>z_j\text{, for all }j\neq k \}\subset\mathbb{R}^K,\text{ }1\leq k\leq K,
\end{equation}
and $\mathcal{R}_0$ is the union of boundaries between $\mathcal{R}_k$’s. Clearly, the network $f_w$ predicts for $x\in\Omega$ a unique label, say, $y_w(x)=\{k\}$, once $f_w(x)\in\mathcal{R}_k$; otherwise, i.e., $\text{card}(y_w(x))>1$, whenever $f_w(x)\in\mathcal{R}_0$. Therefore, the input subset of $\Omega$ over which $y_w(x)=\{k\}$ is
\begin{equation}
\label{eq:3_3}
\Omega_k(w)\triangleq f_w^{-1}(\mathcal{R}_k)=\{x\in\Omega\mid f_w(x)\in\mathcal{R}_k \}, \ 1\leq k\leq K,
\end{equation}
where $f_w^{-1}(\mathcal{R}_k)$ stands for the inverse image of $\mathcal{R}_k$ under $f_w$; all those with $\text{card}(y_w(x))>1$ then belong to $\Omega_0(w)=f_w^{-1}(\mathcal{R}_0)$. The core idea behind is: If two networks $f_w$ and $f_{w'}$ perform close, the event “$\Omega_k(w)$ differs from $\Omega_k(w')$ throughout $0\leq k \leq K$” is of a small probability. To formalize matters, we consider the symmetric difference \cite{Bogachev07} between $\Omega_k(w)$ and $\Omega_k(w')$, namely,
\begin{table}[t!]
\footnotesize
\caption{Pair-wise distances $d_{\mu}(\cdot,\cdot)$ between $\widetilde{w}_1$, $\widetilde{w}_2$, and $\widetilde{w}_3$, with input data obeys truncated Gaussian and uniform distributions, respectively.}
  \vspace{-1.7em}
\label{tab:dis}
\begin{center}
  \begin{tabular}{|c|c|c|c|} \hline
   \diagbox{Distribution}{$d_{\mu}(\cdot,\cdot)$}  & $d_{\mu}(\widetilde{w}_1,\widetilde{w}_2)$ & $d_{\mu}(\widetilde{w}_1,\widetilde{w}_3)$ & $d_{\mu}(\widetilde{w}_2,\widetilde{w}_3)$  \\ \hline
   Truncated Gaussian & 0.8513 & 0.1395 & 0.8482 \\ \hline
   Uniform            & 0.9118 & 0.0828 & 0.9196 \\ \hline
  \end{tabular}
\end{center}
\end{table}
\begin{equation}
\label{eq:3_4}
\Omega_k(w)\Delta\Omega_k(w')\triangleq (\Omega_k(w)\cap \Omega_k(w')^c)\cup(\Omega_k(w)^c\cap\Omega_k(w')),
\end{equation}
and then propose the following distance measure $d_{\mu}(\cdot,\cdot):\mathcal{W}\times\mathcal{W}\to\mathbb{R}^+_0$
\begin{equation}
\label{eq:3_5}
d_{\mu}(w,w')\triangleq \frac{1}{2}\sum_{k=0}^{K}\mu(\Omega_k(w)\Delta\Omega_k(w'))\text{, }w,w'\in\mathcal{W},
\end{equation}
where the scalar $1/2$ is included as a normalization factor. For the three network parameter vectors in Table \ref{tab:toy}, we compare in Table \ref{tab:dis} their $d_{\mu}(\cdot,\cdot)$ for input data sampled from the square region $\Omega=[-3,3]\times[-3,3]$ according to the truncated Gaussian (zero-mean and unit variance) and uniform distributions, respectively. The results show that $d_{\mu}(\widetilde{w}_1,\widetilde{w}_2)$ is large, accounting for their drastically different classification outcomes seen in Fig 2, and $d_{\mu}(\widetilde{w}_1,\widetilde{w}_3)$ is small, well justifying $\widetilde{w}_1$ and $\widetilde{w}_3$ perform close. This example bespeaks that, compared to the Euclidean metric (see Table \ref{tab:toy}), our distance measure \eqref{eq:3_5} can better discriminate networks according to their classification outcomes.

\subsection{Properties of $d_{\mu}(\cdot,\cdot)$}
The distance measure \eqref{eq:3_5} enjoys an appealing property, stated below.

\begin{property}\label{prop:pro_1}
The following equation holds
\begin{equation}
\label{eq:3_6}
d_{\mu}(w,w')=\mu(\{x\in\Omega\mid y_w(x)\neq y_{w'}(x) \})\text{, }w,w'\in\mathcal{W}.
\end{equation}
\begin{proof}
See Appendix \ref{app.A}.
\end{proof}
\end{property}
Hence, $d_{\mu}(w,w')$ gauges how likely the decisions of $f_w$ and $f_{w'}$ disagree. In particular, if $f_w$ is the ground truth classifier, $d_{\mu}(w,w')$ is exactly the \textit{generalization error} \cite{Mohri18,Shwartz14} achieved by $f_{w'}$. In view of this, a network $f_{w'}$ admitting a generalization error less than $\epsilon$ lies in the $\epsilon$-neighborhood centered at the ground truth $f_w$, as illustrated in Fig. 4. With such a nice geometry in mind, we embark on investigating the mathematical structure behind $d_{\mu}(\cdot,\cdot)$, say, if it is a metric \cite{Marsden93,Rudin23} over $\mathcal{W}$. 

\begin{figure}[t!]
  \centering
  \label{fig_4}\includegraphics[width=0.4\columnwidth]{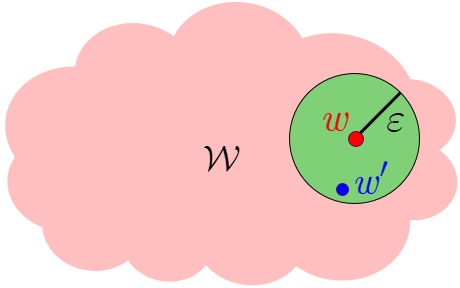}
    \vspace{-1.7em}
  \caption{Schematic depiction of a DL network $f_{w'}$ achieving a generalization error less than $\epsilon$: it lies in the $\epsilon$-neighborhood centered at the ground truth classifier $f_w$.}
\end{figure}

From (3.5), $d_{\mu}$ is for sure non-negative; also, it is symmetric and satisfies the triangular inequality, as asserted below.
\begin{property}\label{prop:pro_2}
For $w,w',w''\in\mathcal{W}$, the following results hold
\begin{enumerate}[label=(\alph*)]
    \item $0\leq d_{\mu}(w,w')=d_{\mu}(w',w).$
    \item $d_{\mu}(w,w')\leq d_{\mu}(w,w'')+d_{\mu}(w'',w').$
\end{enumerate}

\begin{proof}
See Appendix \ref{app.B}.
\end{proof}
\end{property}
However, it is easy to see from \eqref{eq:3_5} that $d_{\mu}(w,w')=0$, i.e., the event “$f_w$ agrees with $f_{w'}$” is of zero probability, does not guarantee $w=w'$, meaning that $d_{\mu}(\cdot,\cdot)$ is not a metric. Fortunately, the story does not just end here, as we elaborate more next.

\section{Metric topology}
\label{sec_4}
\subsection{Metric topology on quotient set of $\mathcal{W}$}
We observe that $d_{\mu}(w,w')=0$ defines an equivalent relation, hereafter denoted by $\cong$, on $\mathcal{W}$. Indeed, reflexivity, i.e., $d_{\mu}(w,w)=0$, $\forall w\in\mathcal{W}$, and symmetry, i.e., $d_{\mu}(w,w')=0$ implies $d_{\mu}(w',w)=0$, follow immediately from definition \eqref{eq:3_5}. Transitivity, that is,  $d_{\mu}(w,w')=d_{\mu}(w',w'')=0$ leads to $d_{\mu}(w,w'')=0$, is a direct consequence of the triangular inequality in Property \ref{prop:pro_2} (b). For $w\in\mathcal{W}$, let
\begin{equation}
\label{eq:4_1}
[w]\triangleq \{w'\in\mathcal{W}\mid d_{\mu}(w',w)=0\}
\end{equation}
be the class consisting of all elements equivalent to $w$, that is to say, of those networks performing as good as $w$ when judged according to the proposed distance measure \eqref{eq:3_5}. In \eqref{eq:4_1}, $w$ is the representation of the class. We then consider the \textit{quotient set of} $\mathcal{W}$ \textit{modulus the equivalent relation} $\cong$, namely,
\begin{equation}
\label{eq:4_2}
\mathcal{W} \slash \cong \triangleq \{[w]\mid w\in\mathcal{W} \}.
\end{equation}
Notably, the union of all equivalent classes in $\mathcal{W} \slash \cong$ forms a disjoint partition of $\mathcal{W}$ \cite{Gamelin99}. We prefer $\mathcal{W} \slash \cong$ to $\mathcal{W}$ because, in $\mathcal{W} \slash \cong$, each $[w]$ is deemed as a “point” to facilitate analyses (say, of performance gaps between distinct classes) on an elegant point-wise basis. On $\mathcal{W} \slash \cong$ we go on to define $d(\cdot,\cdot):\mathcal{W} \slash \cong\times \mathcal{W} \slash \cong\to\mathbb{R}^+_0$ by
\begin{equation}
\label{eq:4_3}
d([w],[w'])\triangleq \frac{1}{2}\sum_{k=0}^{K}\mu(\Omega_k(w)\Delta\Omega_k(w')),
\end{equation}
in order to evaluate the distance between two equivalent classes $[w]\neq [w']$. From \eqref{eq:3_5} and \eqref{eq:4_3}, it is worthy of noting $d([w],[w'])=d_{\mu}(w,w')$, which is exactly the distance between the representations $w$ and $w'$ of the two classes. Even though $d_{\mu}(\cdot,\cdot)$ is not a metric on $\mathcal{W}$, things are totally different if we instead look at the quotient set $\mathcal{W} \slash \cong$ in conjunction with $d(\cdot,\cdot)$. To be specific, the next theorem holds.

\begin{theorem}\label{thm:metric}
The distance measure $d(\cdot,\cdot)$ in \eqref{eq:4_3} is a metric on the quotient set $\mathcal{W} \slash \cong$.
\begin{proof}
Non-negativity, symmetry, and the triangle inequality hold immediately by using Property \ref{prop:pro_2}. It thus remains to prove $d([w'],[w''])=0$ if and only if $[w']=[w'']$. Assuming $[w']=[w'']$, we must have $w''\in[w']$, implying $d([w'],[w''])=d_{\mu}(w',w'')=0$. Conversely, if $[w']\neq[w'']$, then $w'$ and $w''$ yield different classification performances so that $d([w'],[w''])=d_{\mu}(w',w'')\neq0$. The proof is completed.
\end{proof}
\end{theorem}
Hence, $(\mathcal{W} \slash \cong,d)$ is a metric space of all equivalent classes of $\mathcal{W}$, each one composed of networks performing equally well. The distinctive features of our proposal are three-fold. First, while networks performing equally well are “spread over” $\mathcal{W}$ when viewed from the Euclidean metric, they are identified as an equivalent class, thus a unity as a whole, under the proposed equivalent relation. Second, the endowed metric $d(\cdot,\cdot)$ on the associated quotient set $\mathcal{W} \slash \cong$ allows us to accurately assess the performance gap between distinct classes on a point-wise basis (see Fig. 5 for an illustration). Third, capitalized on the rich metric space theory, we stand to investigate some important properties about $(\mathcal{W} \slash \cong,d)$; this is done in the next Section. Before that, we revisit the foregoing toy example in Introduction to illustrate the proposed metric topology.

\begin{figure}[t!]
  \centering
  \label{fig_5}\includegraphics[width=0.95\columnwidth]{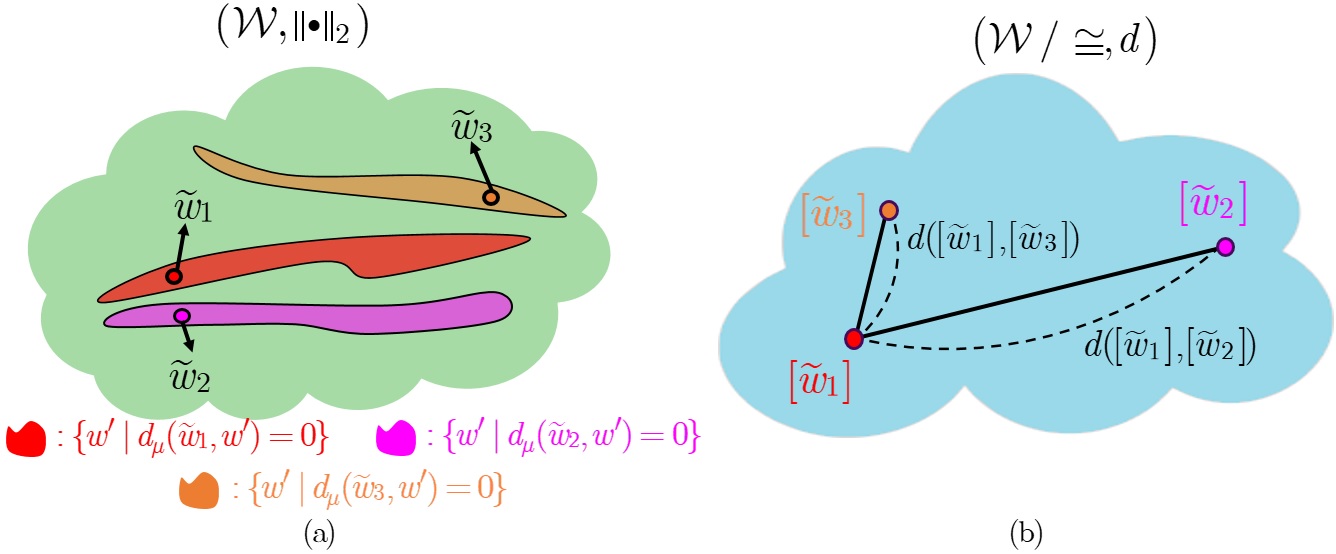}
  \vspace{-1.7em}
  \caption{An illustration of the proposed metric space $(\mathcal{W} \slash \cong,d)$. (a) With the Euclidean metric, DL networks yielding identical classification performance are spread over the space $\mathcal{W}$. (b) They are identified as the same equivalent class in the quotient set $\mathcal{W} \slash \cong$, and the performance gap between distinct classes is assessed by the proposed metric $d$.}
\end{figure}

\begin{figure}[t!]
  \centering
  \label{fig_6}\includegraphics[width=1\columnwidth]{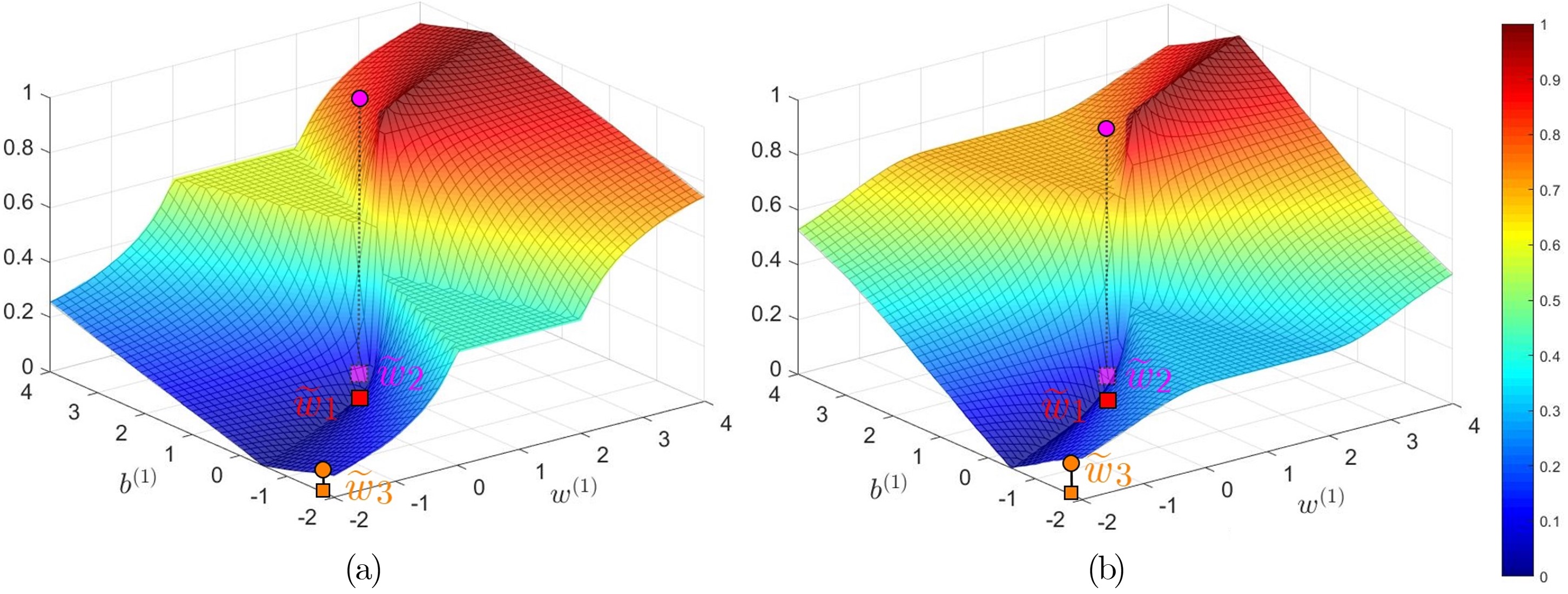}
  \vspace{-2em}
  \caption{An illustration of $d_{\mu}(\widetilde{w}_1,w)$ for $w\in\{w\mid -2\leq w^{(1)}\leq 4, -2\leq b^{(1)}\leq 4 \}$, where $\widetilde{w}_1=(0.8,1,1,1,0.9,1)$. (a) Input data obeys truncated Gaussian distribution with zero mean and identity covariance matrix. (b) Input data obeys uniform distribution.}
\end{figure}
\begin{figure}[t!]
  \centering
  \label{fig_7}\includegraphics[width=0.97\columnwidth]{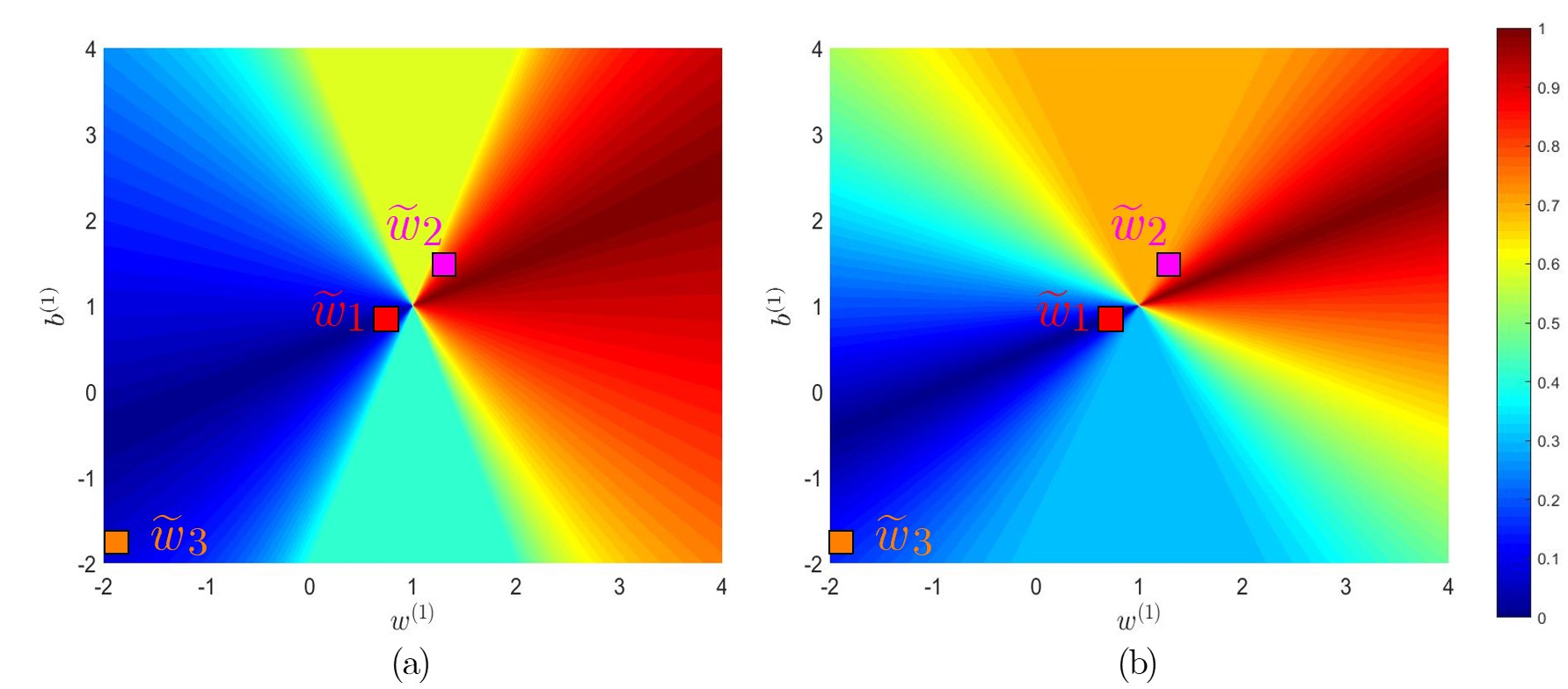}
  \vspace{-1.7em}
  \caption{A bird's eye view illustration of $d_{\mu}(\widetilde{w}_1,w)$ for $w\in\{w\mid -2\leq w^{(1)}\leq 4, -2\leq b^{(1)}\leq 4 \}$, where $\widetilde{w}_1=(0.8,1,1,1,0.9,1)$. (a) Input data obeys truncated Gaussian distribution with zero mean and identity covariance matrix. (b) Input data obeys uniform distribution.}
\end{figure}

\subsection{Example}
\label{sec:4_2}
For the one-layer DL network with six real parameters shown in Fig. 1, it is impossible to directly depict the equivalent classes in the network parameter vector space $\mathcal{W}\subset\mathbb{R}^6$. To ease illustration, we simply set $w^{(2)}=w^{(3)}=w^{(4)}=b^{(2)}=1$ so that only $w^{(1)}$ and $b^{(1)}$ remain free parameters; that is, we just look at the projections onto the two-dimensional $w^{(1)}b^{(1)}$-plane subject to the designated equality constraints. For $\widetilde{w}_1$ given in Table \ref{tab:toy}, Fig. 6-(a) and Fig. 6-(b) plot the computed $d_{\mu}(\widetilde{w}_1,\cdot)$ with respect to the pairs $(w^{(1)},b^{(1)})\in[-2,4]\times[-2,4]$; the input data, drawn from $[-3,3]\times[-3,3]\subset\mathbb{R}^2$, obey the two distributions considered above. In both figures, the pairs $(w^{(1)},b^{(1)})$ achieving zero distance (the darkest color) belong to $[\widetilde{w}_1]$. Notably, there is a sharp change in $d_{\mu}(\widetilde{w}_1,\cdot)$ around $(w^{(1)},b^{(1)})=(1,1)$. Actually, further zooming in on the plot shows the existence of a discontinuity at $(1,1)$, which corresponds to the network parameter vector $(w^{(1)},w^{(2)},w^{(3)},w^{(4)},b^{(1)},b^{(2)})=(1,1,1,1,1,1)$. Such an “all-one” network results in identical network outputs ($f^1_w=f^2_w$ in Fig. 1), then making a random guess between the two indexes. Here, a pair $(w^{(1)},b^{(1)})$ other than $(1,1)$ represents a network certain to predict a unique label. Consequently, the vicinity of $(1,1)$ abounds in such ``normal" networks from different equivalent classes, one with different distance $d_{\mu}(\widetilde{w}_1,\cdot)$: this accounts for the drastic change in $d_{\mu}(\widetilde{w}_1,\cdot)$  around $(1,1)$. A rigorous study of this phenomenon is addressed in the next section. To better visualize the “projected” equivalent classes, Fig. 7-(a) and Fig. 7-(b) plot the “bird’s eye view” of the level surfaces in Fig. 6-(a) and Fig. 6-(b), respectively. The pairs $(w^{(1)},b^{(1)})$ in the same equivalent class (of the same color) are seen to form a ray emanating from $(1,1)$. The figures also clearly show the discontinuity $(1,1)$ is surrounded by elements from distinct equivalent classes.

\section{Theoretical results}\label{sec_5}
Now we are in a position to state the main theoretical results of this paper. The analyses throughout are built on the following assumptions.
\begin{assumption}\label{as:2}
The input domain $\Omega$ is bounded, say,
\begin{equation}
\label{eq:5_1}
\Omega=B(M_1)\triangleq\{x\mid \|x \|_2\leq M_1 \}\subset\mathbb{R}^{n_0}
\end{equation}
for some finite $M_1>0$. 
\end{assumption}

\begin{assumption}\label{as:3}
The set $\mathcal{W}$ of network parameter vectors is bounded, i.e.,
\begin{equation}
\label{eq:5_2}
\mathcal{W}=B(M_2)\triangleq \{w\mid \|w\|_2\leq M_2 \}\subset\mathbb{R}^m
\end{equation}
for some finite $M_2>0$. 
\end{assumption}
Boundedness of the input domain $\Omega$ is valid since practical data samples are of finite power; so is the boundedness of $\mathcal{W}$ because real-world implementations prevent unlimited growth in the total network power. Without loss of generality, we therefore identify the input domain $\Omega$, and the network weight set $\mathcal{W}$ as well, with a bounded and closed Euclidean two-norm ball so that both $\Omega$ and $\mathcal{W}$ are compact in the Euclidean metric; such a property can facilitate our on-going analytic study. Alongside the boundedness assumptions, the following condition to balance the probability measure of event subsets is required.
\begin{assumption}\label{as:4}
The probability measure of each event $U\in\mathcal{A}$ satisfies $\mu(U)\leq \kappa |U|$ for some $\kappa>0$. 
\end{assumption}
Assumption \ref{as:4} excludes impulse-type probability densities; hence small (in Lebesgue measure) events are rare and, in particular, measure-zero events never occur. Such a requirement is fulfilled by many practical data distributions. Indeed, for the input domain $\Omega$ in \eqref{eq:5_1}, we have $\kappa=\Gamma(n_0/2+1)/(\pi^{n_0/2}M^{n_0}_1)$, where $\Gamma(\cdot)$ being the Euler’s Gamma function \cite{Li11}, for the uniform distribution, and $\kappa=1/[\int_{B(M_1)}\exp (-\|x\|^2_2 /2)dx  ]$ for the truncated Gaussian distribution with zero mean and identity covariance matrix.

\subsection{Projection map}
In regards to equivalent relation and the associated quotient set, the projection map \cite{Gamelin99}, specifically in our case $\Pi:\mathcal{W}\to\mathcal{W} \slash \cong$ defined by
\begin{equation}
\label{eq:5_3}
\Pi(w)=[w]\text{, }\forall w\in\mathcal{W},
\end{equation}
is rather pivotal. Here we stress that $\Pi$ is a mapping between two metric spaces, from the Euclidean $\mathcal{W}$ into the quotient set $\mathcal{W} \slash \cong$ endowed with the proposed metric \eqref{eq:4_3}. To proceed, we first recall from Section \ref{sec:3_1} that, for a given $w\in\mathcal{W}$, $\Omega_0(w)=f^{-1}_w(\mathcal{R}_0)$ is the associated input subset such that $\text{card}(y_w(x))>1$. If $\mu(\Omega_0(w))>0$, the network $f_w$ then has a chance to predict for $\mu$-distributed data non-unique labels. The set of such networks is virtually nil in the domain $\mathcal{W}$ of $\Pi$, as asserted by the following theorem.
\begin{theorem}\label{thm:measure_zero}
Define
\vspace{-.5em}
\begin{equation}
\label{eq:5_4}
\mathcal{W}_0\triangleq \{w\in\mathcal{W}\mid\mu(\Omega_0(w))>0 \}\subset \mathcal{W}.
\end{equation}
Under Assumption \ref{as:4}, we have $|\mathcal{W}_0|=0$.                                 \end{theorem}
\vspace{-.5em}
\begin{proof}
See Appendix \ref{app.C}.
\end{proof}
On account of Theorem \ref{thm:measure_zero}, networks certain to predict for $\mu$-distributed data a unique label are almost everywhere in $\mathcal{W}$. In the example given in Section \ref{sec:4_2}, $\mathcal{W}_0$ is a singleton of the all-one vector $(1,1,1,1,1,1)$, corresponding to $(w^{(1)},b^{(1)})=(1,1)$ in the Fig. 6 and 7; it is clearly of Lebesgue measure zero. The projection map $\Pi$ in \eqref{eq:5_3} enjoys the following nice property.
\begin{theorem}\label{thm:continuous}
Under Assumptions \ref{as:measure}$\sim$\ref{as:4}, the projection map $\Pi$ is continuous on $\mathcal{W}$ except for $\mathcal{W}_0$.   
\end{theorem}
\begin{proof}
See Appendix \ref{app.D}.
\end{proof}
\vspace{-.2em}
Discontinuities of $\Pi$ on the subset $\mathcal{W}_0$ is not unexpected. This is because, for $w\in\mathcal{W}_0$, hence $\mu(\Omega_0(w))>0$, there exists in every its neighborhood a $w'\notin\mathcal{W}_0$, with $\mu(\Omega_0(w'))=0$, such that $d([w],[w'])$ is bounded from below by a small factor of $\mu(\Omega_0(w))$, no matter how close $w$ and $w'$ are. As an illustration using the example in Section \ref{sec:4_2}, the discontinuity occurs at $\mathcal{W}_0=\{(1,1,1,1,1,1)\}$, aside from which the distance $d([\widetilde{w}_1],\cdot)=d_{\mu}(\widetilde{w}_1,\cdot)$ varies smoothly. Continuity of $\Pi$ specified by the above theorem will be exploited to devise two important properties regarding the proposed metric topology, as done next.
\subsection{Properties of metric topology $(\mathcal{W} \slash \cong,d)$}
First of all, we note that a subset of Lebesgue measure zero warrants the existence of an arbitrarily small open cover \cite{Marsden93}. Therefore, given $\epsilon>0$ there exists an Euclidean open $S_{\epsilon}\subset\mathcal{W}$ with $|S_{\epsilon}|<\epsilon$ such that $\mathcal{W}_0\subset S_{\epsilon}$. From now on we will call $\mathcal{W}_\epsilon\triangleq \mathcal{W} \backslash S_{\epsilon}$ an $\epsilon$-\textit{pruned} network set, to retain all networks in $\mathcal{W}$ except those in a “vanishingly small” open cover of $\mathcal{W}_0$. When restricted to $\mathcal{W}_\epsilon$, the resultant quotient set endowed with the proposed metric topology is compact. More precisely we have the following.
\vspace{-0.2em}
\begin{theorem}\label{thm:compact}
Under Assumptions \ref{as:measure}$\sim$\ref{as:4}, and given an $\epsilon$-pruned $\mathcal{W}_\epsilon$, the metric space $(\mathcal{W}_{\epsilon} \slash \cong,d)$ is compact.
\begin{proof}
Since $\mathcal{W}_0\subset S_{\epsilon}$, it follows $\mathcal{W}_{\epsilon}\subset\mathcal{W}\backslash\mathcal{W}_0$, and consequently Theorem \ref{thm:continuous} implies the projection map $\Pi$ is continuous on $\mathcal{W}_{\epsilon}$. Moreover, as $\mathcal{W}$ is compact (Assumption \ref{as:3}) and $S_{\epsilon}$ is open, $\mathcal{W}_{\epsilon}=\mathcal{W}\backslash S_{\epsilon}$ remains compact as well. The assertion follows since a continuous mapping preserve compactness \cite{Marsden93}.   
\end{proof}
\end{theorem}
\vspace{-0.2em}
In the literature of topological space, one natural and widely-considered topology on the quotient set $\mathcal{W} \slash \cong$ modulo the equivalent relation $\cong$ is the quotient topology \cite{Gamelin99}, which is the collection of all subsets $U\subset\mathcal{W} \slash \cong$ such that the inverse image $\Pi^{-1}(U)$ under the projection map $\Pi$ is open in $\mathcal{W}$. The following theorem asserts that, if we choose to stay in $\mathcal{W}_{\epsilon}$, our proposed metric topology on $(\mathcal{W}_{\epsilon} \slash \cong,d)$ coincides with the quotient topology.
\begin{theorem}\label{thm:quotient}
Under Assumptions \ref{as:measure}$\sim$\ref{as:4}, and given an $\epsilon$-pruned $\mathcal{W}_\epsilon$, the collection $\mathcal{T}$ of all open subsets in the metric space $(\mathcal{W}_{\epsilon} \slash \cong,d)$ is the quotient topology on $\mathcal{W}_{\epsilon} \slash \cong$.  
\begin{proof}
It suffices to show that
\begin{enumerate}[label=(\alph*)]
    \item For any $T\in\mathcal{T}$, i.e., $T$ is open in $(\mathcal{W}_{\epsilon} \slash \cong,d)$, the inverse image $\Pi^{-1}(T)$ is open in $\mathcal{W}_{\epsilon}$.
    \item If $V$ is an open subset such that $V=\Pi^{-1}(T)$ for some $T\in\mathcal{W}_{\epsilon} \slash \cong$, then $T$ lies in $\mathcal{T}$.
\end{enumerate}
Part (a) is clearly true thanks to the continuity of the projection $\Pi$ shown in Theorem \ref{thm:continuous}. We then go on to prove part (b). Since $V^c=\mathcal{W}_{\epsilon} \backslash V$ is a closed subset of a bounded set $\mathcal{W}_{\epsilon}$ in the Euclidean space, by the Heine–Borel theorem \cite{Rudin23} $V^c$ is compact. Since the projection $\Pi$ is continuous, $\Pi(V^c)=(\mathcal{W}_{\epsilon} \slash \cong)\backslash T=T^c$ is compact \cite[Theorem 4.2.2]{Marsden93} in $(\mathcal{W}_{\epsilon} \slash \cong,d)$. Since every compact set in a metric space is closed \cite[p.174]{Marsden93}, $T$ is open in $(\mathcal{W}_{\epsilon} \slash \cong,d)$. The proof is thus completed. 
\end{proof}
\end{theorem}

\section{Summary and discussions}\label{sec_6}
The less than tractable nonlinearity and layered architecture have posed a big challenge for theoretical analyses and study of DL. While researchers have raced for developing mathematical foundations of DL, current achievements would seem far from complete and unified. So new mathematical elements able to gain further insights into DL are definitely important and welcome. This paper presents some of our findings through the lens of metric topology, a fundamental and well-established tool in mathematical analysis yet still absent from the DL literature. We first propose a probabilistic-based distance measure, as a replacement of conventional Euclidean metric, to assess the performance gap between two networks according to their classification performances. This distance measure is reminiscent of the thread in statistical learning theory, specialized to the well-known generalization error of a network when compared against the ground truth classifier. Although not a metric among network parameter vectors, it defines an equivalent relation able to identify networks performing equally well with one single equivalent class. Our distance measure then provably serves as a metric on the quotient set modulo the equivalent relation. Under quite mild conditions, we go on to present a series of analyses into the proposed metric space. Continuity of the projection map, except for a Lebesgue measure zero subset of networks likely to mark ambiguous labels, is first established. Aided by this result, two important properties regarding our metric topology, namely, compactness and equivalence to the well-known quotient topology, are then established. 

Our study opens up several interesting future research directions. First and foremost, how the proposed quotient set and metric space formulation can shed light on algorithmic design is of high interest. Since compactness implies completeness, one possible approach is devising new design objectives that are contraction \cite{Marsden93,Rudin23}; the celebrated Banach contraction mapping principle \cite{Agarwal18,Gohberg81} along with simple fixed-point iteration then offers a potential framework for algorithm design. Of further topological aspects yet to be explored is connectivity, an enabling property for developing algorithms on the basis of algebraic topology \cite{Hatcher02,Spanier66,Strang19}. Although the metric topology proposed in this work is specifically constructed and applied to data classification problems, it can be extended to deal with DL-empowered regression and multi-armed bandit problems \cite{Xu22,Zhu21}
according to our preliminary investigation.

\begin{appendices}\vspace{-0.5em}
\section{Proof of Property \ref{prop:pro_1}} 
\label{app.A}
\vspace{-0.5em}
Starting from \eqref{eq:3_5}, we have
\begin{equation}
\begin{aligned}
\label{eq:A_1}
&d_{\mu}(w,w')=\frac{1}{2}\sum_{k=0}^K\mu\left(\Omega_k(w)\Delta\Omega_k(w')\right) 
\\ &  =\frac{1}{2}\sum_{k=0}^K \mu \big(\left( \Omega_k(w)\cap\Omega_k(w')^c \right) \cup \left( \Omega_k(w)^c\cap\Omega_k(w') \right)   \big)
\\ & =\frac{1}{2}\sum_{k=0}^K \Big[\mu \big( \Omega_k(w)\cap\Omega_k(w')^c \big) + \mu\big(  \Omega_k(w)^c\cap\Omega_k(w') \big)\Big]
\\& \stackrel{(a)}{=} \frac{1}{2}\Big[\mu \big( \Omega_0(w)\cap\Omega_0(w')^c \big) + \mu\big(  \Omega_0(w)^c\cap\Omega_0(w') \big)\Big]
\\ &\quad +\frac{1}{2}\sum_{k=1}^K \Big[\mu \big( \{x\in\Omega\mid f_{w}(x)\in \mathcal{R}_k, f_{w'}(x)\notin \mathcal{R}_k   \} \big) + \mu\big(  \{x\in\Omega\mid f_{w}(x)\notin \mathcal{R}_k, f_{w'}(x)\in \mathcal{R}_k   \} \big)\Big]
\\ & \stackrel{(b)}{=} \frac{1}{2}\Big[\mu \big( \{x\in\Omega\mid \text{card}(y_{w}(x))>1, \text{card}(y_{w'}(x))=1  \}\big) + \mu \big( \{x\in\Omega\mid \text{card}(y_{w'}(x))>1, \text{card}(y_{w}(x))=1  \}\big)\Big]
\\ &\quad +\frac{1}{2}\sum_{k=1}^K \Big[\mu \big( \{x\in\Omega\mid y_{w}(x)=k, y_{w'}(x)\neq k   \} \big) + \mu\big(  \{x\in\Omega\mid y_{w'}(x)=k, y_{w}(x)\neq k      \} \big)\Big]
\\ & = \frac{1}{2}\Big[\mu \big( \{x\in\Omega\mid \text{card}(y_{w}(x))>1, y_{w'}(x)\neq y_{w}(x)  \}\big) + \mu \big( \{x\in\Omega\mid \text{card}(y_{w'}(x))>1, y_{w}(x)\neq y_{w'}(x)  \}\big)\Big]
\\ &\quad +\frac{1}{2}\sum_{k=1}^K \Big[\mu \big( \{x\in\Omega\mid y_{w}(x)=k, y_{w'}(x)\!\neq\! y_{w}(x)   \} \big) \!+\! \mu\big(  \{x\in\Omega\mid y_{w'}(x)\!=\!k, y_{w}(x)\neq y_{w'}(x)  \} \big)\Big]
\\ & = \frac{1}{2}\Big[\mu \big( \{x\in\Omega_0(w)\mid y_{w'}(x)\neq y_{w}(x)\}\big) + \mu \big( \{x\in\Omega_0(w')\mid y_{w}(x)\neq y_{w'}(x)  \}\big)\Big]
\\ &\quad +\frac{1}{2}\sum_{k=1}^K \Big[\mu \big( \{x\in\Omega_k(w)\mid y_{w'}(x)\neq y_{w}(x)\}\big) + \mu \big( \{x\in\Omega_k(w')\mid y_{w}(x)\neq y_{w'}(x)  \}\big)\Big]
\\& \stackrel{(c)}{=} \frac{1}{2}\Big[\mu \big( \{x\in\Omega\mid y_{w'}(x)\neq y_{w}(x)\}\big) + \mu \big( \{x\in\Omega\mid y_{w}(x)\neq y_{w'}(x)  \}\big)\Big]
\\ & = \mu \big( \{x\in\Omega\mid y_{w'}(x)\neq y_{w}(x)\}\big),
\end{aligned}
\end{equation}
where (a), (b), and (c) follow from \eqref{eq:3_3}, \eqref{eq:3_2}, and \eqref{eq:3_1}, respectively. The proof is thus completed. 

\section{Proof of Property \ref{prop:pro_2}} 
\label{app.B}
\vspace{-0.5em}
Part (a) holds directly holds from \eqref{eq:3_4} and \eqref{eq:3_5} since
\begin{equation}
\begin{aligned}
\label{eq:B_1}
\quad d_{\mu}(w,w')&=\frac{1}{2}\sum_{k=0}^K\mu\left(\Omega_k(w)\Delta\Omega_k(w')\right)
\\ & =\frac{1}{2}\sum_{k=0}^K \mu \big(\left( \Omega_k(w)\cap\Omega_k(w')^c \right) \cup \left( \Omega_k(w)^c\cap\Omega_k(w') \right)   \big)
\\ & =\frac{1}{2}\sum_{k=0}^K \mu \big(\left( \Omega_k(w')\cap\Omega_k(w)^c \right) \cup \left( \Omega_k(w')^c\cap\Omega_k(w) \right)   \big)=d_{\mu}(w',w).
\end{aligned}
\end{equation}
To prove part (b), let $x\in\Omega_k(w)\Delta\Omega_k(w')$, therefore either $x\in\Omega_k(w)\cap\Omega_k(w')^c$ or $x\in\Omega_k(w)^c\cap\Omega_k(w')$. For the case that $x\in\Omega_k(w)\cap\Omega_k(w')^c$, either $x\in\big(\Omega_k(w)\cap\Omega_k(w')^c\big)$ $\cap \  \Omega_k(w'')\subset\big(\Omega_k(w'')\cap \Omega_k(w')^c \big)$ or $x\in\big(\Omega_k(w)\cap\Omega_k(w')^c\big)\cap\Omega_k(w'')^c\subset\big(\Omega_k(w)\cap \Omega_k(w'')^c \big)$, which implies that
\begin{equation}
\begin{aligned}
\label{eq:B_2}
x\in \big(\Omega_k(w)\cap \Omega_k(w'')^c \big)\cup\big(\Omega_k(w'')\cap \Omega_k(w')^c \big).
\end{aligned}
\end{equation}
Similar to \eqref{eq:B_2}, for the case that $x\in\Omega_k(w)^c\cap\Omega_k(w')$ we have
\begin{equation}
\begin{aligned}
\label{eq:B_3}
x\in \big(\Omega_k(w)^c\cap \Omega_k(w'') \big)\cup\big(\Omega_k(w'')^c\cap \Omega_k(w') \big).
\end{aligned}
\end{equation}
With \eqref{eq:B_2} and \eqref{eq:B_3}, it follows that
\begin{equation}
\begin{aligned}
\label{eq:B_4}
&\Omega_k(w)\Delta\Omega_k(w')
\\&\subset   \Big[ \big(\Omega_k(w) \cap  \Omega_k(w'')^c \big) \cup \big(\Omega_k( w'') \cap  \Omega_k(w')^c \big) \Big] \cup  \Big[ \big(\Omega_k(w)^c \cap  \Omega_k(w'') \big) \cup \big(\Omega_k( w'')^c \cap  \Omega_k(w')  \big) \Big]
\\&=   \Big[ \big(\Omega_k(w) \cap  \Omega_k(w'')^c \big) \cup \big(\Omega_k(w)^c \cap  \Omega_k(w'') \big) \Big] \cup  \Big[ \big(\Omega_k( w'') \cap  \Omega_k(w')^c \big) \cup \big(\Omega_k( w'')^c \cap  \Omega_k(w')  \big) \Big]
\\&=  \Big[\Omega_k(w)\Delta\Omega_k(w'')\Big]\cup\Big[\Omega_k(w'')\Delta\Omega_k(w')\Big],
\end{aligned}
\end{equation}
which in conjunction with the subadditivity of measure yields
\begin{equation}
\begin{aligned}
\label{eq:B_5}
d_{\mu}(w,w')&=\frac{1}{2}\sum_{k=0}^K\mu\left(\Omega_k(w)\Delta\Omega_k(w')\right)
\\&\leq \frac{1}{2}\sum_{k=0}^K\mu\left(\Omega_k(w)\Delta\Omega_k(w'')\cup \Omega_k(w'')\Delta\Omega_k(w')\right)
\\&\leq \frac{1}{2}\sum_{k=0}^K\big[\mu\left(\Omega_k(w)\Delta\Omega_k(w'')\right)+\mu\left( \Omega_k(w'')\Delta\Omega_k(w')\right)\big]
\\& =d_{\mu}(w,w'')+d_{\mu}(w'',w').
\end{aligned}
\end{equation}
The proof is thus completed.   

\section{Proof of Theorem \ref{thm:measure_zero}} 
\label{app.C}
By the definition \eqref{eq:3_3}, we have 
\begin{equation}
\begin{aligned}
\label{eq:6_1}
\Omega_0(w)&=\{x\in\Omega\mid\text{card}(y_w(x))>1 \} \\ & \stackrel{(a)}{=}\bigcup_{t=1}^K\bigcup_{t\neq s}\{x\in\Omega\mid \exists 1\leq t\neq s\leq K \text{ s.t. }(f^t_w-f^s_w)(x)=0\},
\end{aligned}
\end{equation}
where (a) follows directly from the decision rule \eqref{eq:2_5}. Using \eqref{eq:6_1} along with the union bound we reach
\begin{equation}
\begin{aligned}
\label{eq:6_2}
|\mathcal{W}_0|&=\big|\{w\in\mathcal{W}\mid\mu(\Omega_0(w))>0  \}\big|\\& \leq \sum_{t=1}^K \sum_{s=1,s\neq t}^K \big|\{w\in\mathcal{W}\mid \mu( \{ x\in\Omega\mid (f^t_w-f^s_w)(x)=0 \} ) >0 \}\big|.
\end{aligned}
\end{equation}
Hence, $|\mathcal{W}_0|=0$ once the right-hand-side of \eqref{eq:6_2} is zero, which is true if 
\begin{equation}
\begin{aligned}
\label{eq:6_3}
\left|\left\{w\in\mathcal{W}\mid \mu\left( \{ x\in\Omega\mid (f^t_w-f^s_w)(x)=0 \} \right) >0 \right\}\right|=0,\text{ } t\neq s.
\end{aligned}
\end{equation}
Under Assumption \ref{as:4}, a sufficient condition for \eqref{eq:6_3} is
\begin{equation}
\begin{aligned}
\label{eq:6_4}
\left|\left\{w\in\mathbb{R}^m\mid (f^t_w-f^s_w)(x)=0 \right\}\right|=0\text{, for all } x\in\mathbb{R}^{n_0}.
\end{aligned}
\end{equation}
Indeed, supposing \eqref{eq:6_4} is true (the proof of \eqref{eq:6_4} is shown later) we immediately have
\begin{equation}
\begin{aligned}
\label{eq:6_5}
\left|\left\{(x,w)\in\mathbb{R}^{n_0+m}\mid (f^t_w-f^s_w)(x)=0 \right\}\right|&=\iint_{\mathbb{R}^{n_0+m}}\mathbf{1}((f^t_w-f^s_w)(x)=0)dxdw \\ &\stackrel{(a)}{=}\int_{\mathbb{R}^{n_0}}\left[\int_{\mathbb{R}^{m}} \mathbf{1}((f^t_w-f^s_w)(x)=0)dw \right]dx
\\ &\stackrel{(b)}{=}\int_{\mathbb{R}^{n_0}}0dx=0,
\end{aligned}
\end{equation}
where $\mathbf{1}(\cdot)$ is the indicator function \cite[p.68]{Wheeden15}, (a) follows from the Tonelli’s theorem \cite[Theorem 6.10]{Wheeden15}, and (b) holds due to \eqref{eq:6_4}. Use the Tonelli’s theorem again to rewrite \eqref{eq:6_5} as
\begin{equation}
\begin{aligned}
\label{eq:6_6}
&\left|\left\{(x,w)\in\mathbb{R}^{n_0+m}\mid (f^t_w-f^s_w)(x)=0 \right\}\right|
\\&=\int_{\mathbb{R}^{m}}\left[\int_{\mathbb{R}^{n_0}} \mathbf{1}((f^t_w-f^s_w)(x)=0)dx \right]dw=0.
\end{aligned}
\end{equation}
Hence, the equality $\int_{\mathbb{R}^{n_0}} \mathbf{1}((f^t_w-f^s_w)(x)=0)dx=0$ holds almost everywhere in $\mathbb{R}^m$. As a result, we have 
\begin{equation}
\begin{aligned}
\label{eq:6_7}
&\left|\left\{w\in\mathcal{W}\mid \left|\left\{ x\in\mathbb{R}^{n_0}\mid (f^t_w-f^s_w)(x)=0 \right\}\right|  >0 \right\}\right|
\\ &=\left|\left\{w\in\mathcal{W}\mid    \int_{\mathbb{R}^{n_0}}\mathbf{1}\left( (f^t_w-f^s_w)(x)=0   \right)dx  >0 \right\}\right|=0.
\end{aligned}
\end{equation}
Since, from Assumption \ref{as:4},
\begin{equation}
\begin{aligned}
\label{eq:6_8}
\mu\left( \left\{ x\in\Omega\mid (f^t_w-f^s_w)(x)=0     \right\}   \right)/\kappa&\leq \left|\left\{x\in\Omega\mid (f^t_w-f^s_w)(x)=0   \right\}\right|
\\& \leq \left|\left\{x\in\mathbb{R}^{n_0}\mid (f^t_w-f^s_w)(x)=0   \right\}\right|,
\end{aligned}
\end{equation}
it follows
\begin{equation}
\begin{aligned}
\label{eq:6_9}
&\left\{w\in\mathcal{W}\mid  \mu\left( \left\{ x\in\Omega\mid (f^t_w-f^s_w)(x)=0     \right\}   \right)>0 \right\}  
\\ &\subset \left\{w\in\mathcal{W}\mid \left|\left\{x\in\mathbb{R}^{n_0}\mid (f^t_w-f^s_w)(x)=0   \right\}\right|   >0 \right\}.
\end{aligned}
\end{equation}
Combining \eqref{eq:6_7} and \eqref{eq:6_9} then proves \eqref{eq:6_3}.

\begin{figure}[t!]
  \centering
  \label{fig_8}\includegraphics[width=0.58\columnwidth]{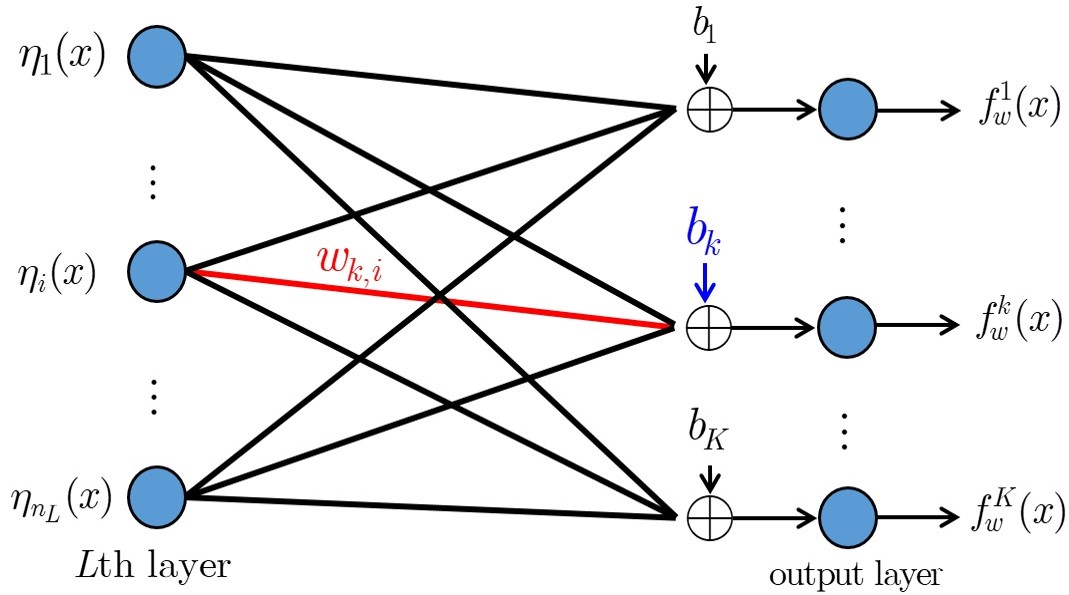}
  \vspace{-1.7em}
  \caption{An illustration of input-output relation between the $L$th hidden layer and the output layer of a DL network.}
\end{figure}

We then go on to prove \eqref{eq:6_4}. Towards this end, let $\eta_i(x)$ be the output of the $i$th node at the $L$th layer, $w_{k,i}$ be the weight of the edge connecting the $i$th node at the $L$th layer and the $k$th node at the output layer, and $b_k$ be the bias of the $k$th node at the output layer; see Fig. 8 for an illustration. From \eqref{eq:2_1}, a compact representation of $(f^t_w-f^s_w)(x)$ is
\begin{equation}
\begin{aligned}
\label{eq:6_10}
(f^t_w-f^s_w)(x)=\sigma\left(\sum_{i=1}^{n_L}w_{t,i}\eta_{i}(x)+b_t  \right)-\sigma\left(\sum_{i=1}^{n_L}w_{s,i}\eta_{i}(x)+b_s  \right).
\end{aligned}
\end{equation}
Since the activation function $\sigma$ is strictly increasing, it is clear that $(f^t_w-f^s_w)(x)=0$ if and only if
\begin{equation}
\begin{aligned}
\label{eq:6_11}
b_s=\sum_{i=1}^{n_L}(w_{t,i}-w_{s,i})\eta_{i}(x)+b_t,
\end{aligned}
\end{equation}
and therefore
\begin{equation}
\begin{aligned}
\label{eq:6_12}
\left\{w\in\mathbb{R}^m \mid (f^t_w-f^s_w)(x)=0 \right\}=\left\{w\in\mathbb{R}^m \mid b_s=\sum_{i=1}^{n_L}(w_{t,i}-w_{s,i})\eta_{i}(x)+b_t \right\}.
\end{aligned}
\end{equation}
Let us stack all the weight and bias parameters except $b_s$ into $\widehat{w}\in\mathbb{R}^{m-1}$. Then we deduce from \eqref{eq:6_12} that
\begin{equation}
\begin{aligned}
\label{eq:6_13}
\left|\left\{w\in\mathbb{R}^m \mid (f^t_w-f^s_w)(x)=0 \right\}\right| &= \left|\left\{w\in\mathbb{R}^m \mid b_s=\sum_{i=1}^{n_L}(w_{t,i}-w_{s,i})\eta_{i}(x)+b_t \right\}\right|
\\&= \int_{\mathbb{R}^m}\mathbf{1}\left( b_s=\sum_{i=1}^{n_L}(w_{t,i}-w_{s,i})\eta_{i}(x)+b_t\right)dw
\\ &\stackrel{(a)}{=}\int_{\mathbb{R}^{m-1}}\int_{\mathbb{R}}\mathbf{1}\left( b_s\!=\!\sum_{i=1}^{n_L}(w_{t,i}-w_{s,i})\eta_{i}(x)+b_t\right)db_sd\widehat{w}
\\ &\stackrel{(b)}{=}\int_{\mathbb{R}^{m-1}}0d\widehat{w}=0,
\end{aligned}
\end{equation}
where (a) follows from the Tonelli’s theorem, and (b) holds since a singleton is measure zero in  $\mathbb{R}$. The proof is thus completed.   

\section{Proof of Theorem \ref{thm:continuous}} 
\label{app.D}
We first prove that the projection map \eqref{eq:5_3} is continuous on $\mathcal{W}\backslash\mathcal{W}_0$. Let $\epsilon>0$ and $w',w''\in\mathcal{W}\backslash\mathcal{W}_0$ such that $\|w'-w''\|_2\leq \epsilon$. Our goal is to show $d([w'],[w''])\to0$ as $\epsilon\to0$. Note that, since $w',w''\notin\mathcal{W}_0$, \eqref{eq:5_4} implies $\mu(\Omega_0(w'))=\mu(\Omega_0(w''))=0$, consequently $\mu(\Omega_0(w')\Delta\Omega_0(w''))=0$. Then, using \eqref{eq:4_3}, it follows
\begin{equation}
\begin{aligned}
\label{eq:6_14}
d([w'],[w''])&=\frac{1}{2}\sum_{k=1}^K\mu\left(\Omega_k(w')\Delta\Omega_k(w'')\right)
\\ & \stackrel{(a)}{=}\frac{1}{2}\sum_{k=1}^K \mu \left(\left( \Omega_k(w')\cap\Omega_k(w'')^c \right) \cup \left( \Omega_k(w')^c\cap\Omega_k(w'') \right)   \right)
\\ & \stackrel{(b)}{=}\frac{1}{2}\sum_{t=1}^K\sum_{s=1,s\neq t}^K \mu \left(\left\{x\in\Omega\mid f_{w'}(x)\in\mathcal{R}_t, f_{w''}(x)\in\mathcal{R}_s \right\}\right)
\\ & \stackrel{(c)}{\leq}\frac{1}{2}\sum_{t=1}^K\sum_{s=1,s\neq t}^K \mu \left(\left\{x\in\Omega\mid (f^t_{w'}-f^s_{w'})(x)>0, (f^t_{w''}-f^s_{w''})(x)<0 \right\}\right),
\end{aligned}
\end{equation}
where (a), (b), and (c) follow from \eqref{eq:3_4}, \eqref{eq:3_3}, and \eqref{eq:3_2}, respectively. Since $\mathcal{W}$ is compact (under Euclidean metric) and $f^k_w$, $1\leq k\leq K$, are continuous functions, $f^k_w$’s are uniformly continuous \cite[Theorem 4.6.2]{Marsden93}. This guarantees for each $1\leq k\leq K$ the existence of a $\delta_k(\epsilon)>0$ such that
\begin{equation}
\begin{aligned}
\label{eq:6_15}
(f^k_{w}-f^k_{\widehat{w}})(x)\in\left(-\delta_k(\epsilon)/2,\delta_k(\epsilon)/2 \right)\text{, }\forall \|w-\widehat{w} \|_2\leq \epsilon.
\end{aligned}
\end{equation}
If we pick $\delta(\epsilon)=\underset{1\leq k\leq K}{\max}\delta_k(\epsilon)$, \eqref{eq:6_15} then implies
\begin{equation}
\begin{aligned}
\label{eq:6_16}
(f^k_{w}-f^k_{\widehat{w}})(x)\in\left(-\delta(\epsilon)/2,\delta(\epsilon)/2 \right)\text{, }\forall \|w-\widehat{w} \|_2\leq \epsilon\text{, for all }1\leq k\leq K,
\end{aligned}
\end{equation}
In particular, for  $w',w''\in\mathcal{W}\backslash\mathcal{W}_0$ with $\|w'-w''\|_2\leq \epsilon$, \eqref{eq:6_16} implies
\begin{equation}
\begin{aligned}
\label{eq:6_17}
f^t_{w''}(x)+\delta(\epsilon)/2>f^t_{w'}(x)\text{ and }f^s_{w'}(x)>f^s_{w''}(x)-\delta(\epsilon)/2.
\end{aligned}
\end{equation}
Accordingly, if $f^t_{w'}(x)>f^s_{w'}(x)$, \eqref{eq:6_17} then gives $f^t_{w''}(x)+\delta(\epsilon)/2>f^s_{w''}(x)-\delta(\epsilon)/2$, thereby
\begin{equation}
\begin{aligned}
\label{eq:6_18}
\{x\in\Omega\mid(f^t_{w'}-f^s_{w'})(x)>0 \}\subset\{x\in\Omega\mid(f^t_{w''}-f^s_{w''})(x)>-\delta(\epsilon) \}.
\end{aligned}
\end{equation}
Using \eqref{eq:6_18}, we immediately have
\begin{equation}
\begin{aligned}
\label{eq:6_19}
&\{x\in\Omega\mid(f^t_{w'}-f^s_{w'})(x)>0, (f^t_{w''}-f^s_{w''})(x)<0 \}\\&\subset\{x\in\Omega\mid(f^t_{w''}-f^s_{w''})(x)\in(-\delta(\epsilon),0 )\}.
\end{aligned}
\end{equation}
Combining \eqref{eq:6_14} and \eqref{eq:6_19} yields
\begin{equation}
\begin{aligned}
\label{eq:6_20}
d([w'],[w''])\leq \frac{1}{2} \sum_{t=1}^K\sum_{s=1,s\neq t}^K \mu \left(\{x\in\Omega\mid(f^t_{w''}-f^s_{w''})(x)\in(-\delta(\epsilon),0 )\}\right).
\end{aligned}
\end{equation}
On the other hand, since $f^t_{w''}$, $1\leq t\leq K$, is continuous over the input domain $\Omega$, associated with a positive integer $q$ the inverse image of $(-1/q,0)$ under $f^t_{w''}-f^s_{w''}$
\begin{equation}
\begin{aligned}
\label{eq:6_21}
E^q_{t,s}= \{x\in\Omega\mid(f^t_{w''}-f^s_{w''})(x)\in(-1/q,0 )\}\subset\Omega,\text{ }1\leq t\neq s\leq K,
\end{aligned}
\end{equation}
is (Euclidean) open, thus measurable. Since $E^q_{t,s}\searrow\{x\in\Omega\mid(f^t_{w''}-f^s_{w''})(x)=0\}$ as $q\to\infty$ and $\mu(E^q_{t,s})\leq1$ is finite, the Bounded Convergence Theorem (e.g., \cite[Theorem 10.11]{Wheeden15}) implies
\begin{equation}
\begin{aligned}
\label{eq:6_22}
\lim_{q\to\infty}\mu(E^q_{t,s})&=\mu\left(\{x\in\Omega\mid(f^t_{w''}-f^s_{w''})(x)=0\} \right)
\\ & \stackrel{(a)}{\leq}\mu(\Omega_0(w''))\stackrel{(b)}{=}0,
\end{aligned}
\end{equation}
where (a) holds by the definition of $\Omega_0(w)$ below \eqref{eq:3_3}, and (b) holds due to $w''\notin\mathcal{W}_0$. Setting $\epsilon\to0$ on both sides of \eqref{eq:6_20} along with \eqref{eq:6_22} then gives $d([w'],[w''])\to0$. Thus, we complete the proof that the projection map \eqref{eq:5_3} is continuous on $\mathcal{W}\backslash\mathcal{W}_0$.
We then go on to show that the projection map \eqref{eq:5_3} is discontinuous on $\mathcal{W}_0$. Let  $\epsilon>0$ and $w'\in\mathcal{W}_0$. Since $\mathcal{W}_0$ is of Lebesgue measure zero, there exists $w''\in\{w\mid\|w-w'\|\leq \epsilon\}$ such that $w''\notin\mathcal{W}_0$ and thus $\mu(\Omega_0(w''))=0$. By the definition \eqref{eq:4_3}, we have
\begin{equation}
\begin{aligned}
\label{eq:6_23}
d([w'],[w''])&=\frac{1}{2}\sum_{k=0}^K\mu\left(\Omega_k(w')\Delta\Omega_k(w'')\right)
\\& \geq \frac{1}{2}\mu\left(\Omega_0(w')\Delta\Omega_0(w'')\right)
\end{aligned}
\end{equation}
\begin{equation}
\begin{aligned}
\notag
&  = \frac{1}{2} \mu \left(\left( \Omega_0(w')\cap\Omega_k(w'')^c \right) \cup \left( \Omega_0(w')^c\cap\Omega_0(w'') \right)   \right)
\\&  \geq \frac{1}{2} \mu \left( \Omega_0(w')\cap\Omega_0(w'')^c \right)   
\\& \geq \frac{1}{2} \mu \left( \Omega_0(w') \right) -\frac{1}{2} \mu \left( \Omega_0(w'') \right)
\\& = \frac{1}{2} \mu \left( \Omega_0(w') \right) \stackrel{(a)}{>}0,
\end{aligned}
\end{equation}
where (a) holds since $w'\in\mathcal{W}_0$. Hence, $d([w'],[w''])$ is lower bounded by $\mu \left( \Omega_0(w') \right)/2$ $>0$, no matter how small $\epsilon$ is. Hence, the projection map \eqref{eq:5_3} is discontinuous on $\mathcal{W}_0$. The proof is thus completed.    
\end{appendices}



\ifCLASSOPTIONcaptionsoff
  \newpage
\fi

\bibliographystyle{IEEEtran}
\bibliography{ref}

\end{document}